\documentclass[11pt,a4paper]{article}
\usepackage[hyperref]{acl2021}
\usepackage{times}
\usepackage{latexsym}

\usepackage{microtype}

\aclfinalcopy %
\def\aclpaperid{713} %

\usepackage{tikz}
\usepackage{amsmath}
\usepackage{amssymb}
\usepackage{amsthm}
\usepackage{booktabs}
\usepackage{enumitem}
\usepackage{multirow}
\usepackage[capitalize]{cleveref}
\newcommand{\comment}[1]{}
\usepackage{inconsolata}
\usepackage{stmaryrd}

\makeatletter
\crefname{section}{\S\@gobble}{\S\S\@gobble}
\crefname{subsection}{\S\@gobble}{\S\S\@gobble}
\makeatother

\usepackage[textsize=footnotesize]{todonotes}

\newcommand{\reals}{\mathbb{R}}

\title{Implicit Representations of Meaning  in Neural Language Models}

\author{Belinda Z.\ Li ~~~ Maxwell Nye ~~~ Jacob Andreas \\
  Massachusetts Institute of Technology \\
  \texttt{\{bzl,mnye,jda\}@mit.edu}
  }

\date{}

\begin{document}
\maketitle

\begin{abstract}
Does the effectiveness of neural language models derive entirely from accurate
modeling of surface word co-occurrence statistics, or do these models
represent and reason about the world they describe? In BART and T5
transformer language models, we identify contextual
word representations that function as \emph{models of entities and situations}
as they evolve throughout a discourse. These neural representations have
functional similarities to linguistic models of dynamic semantics: they
support a linear readout of each entity's current properties and relations, and
can be manipulated with predictable effects on language generation. Our
results indicate that prediction in pretrained neural language models is
supported, at least in part, by dynamic representations of meaning and
implicit simulation of entity state, and that this behavior can be learned 
with only text as training data.\footnote{Code and data are available at
\url{https://github.com/belindal/state-probes}.}
\end{abstract}

\section{Introduction}

Neural language models (NLMs), which place probability distributions over
sequences of words, produce contextual word and sentence embeddings that are useful for a variety of language processing tasks~\citep{peters-etal-2018-deep,lewis-etal-2020-bart}. This usefulness is partially explained by the fact that NLM
representations 
encode lexical relations
 \cite{mikolov-etal-2013-linguistic} and syntactic structure~\citep{tenney-etal-2019-bert}. 
But the extent to which NLM training also induces representations of \emph{meaning} remains a topic of ongoing debate~\citep{bender-koller-2020-climbing,wu2020infusing}.
In this paper, we show that NLMs represent meaning in a specific sense:
in simple semantic domains, they build representations of situations and entities that encode logical descriptions of each entity's dynamic state.

\begin{figure}[t!]
    \centering
    \includegraphics[width=\columnwidth, clip, trim=0 3.9in 7.35in 1.1in]{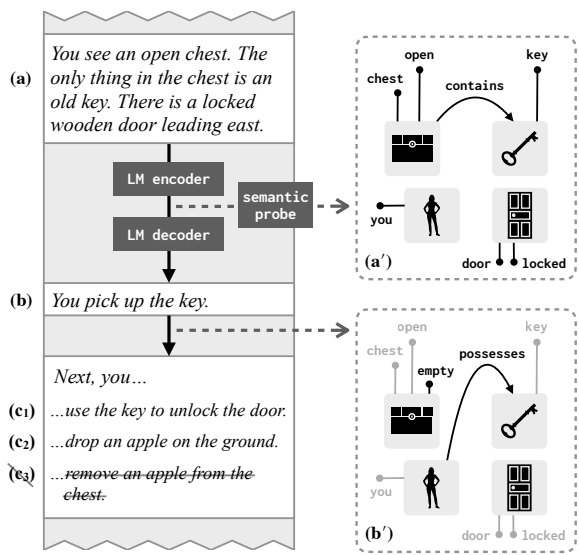}
    \caption{Neural language models trained on text alone (a--c) produce semantic representations that encode properties and relations of entities mentioned in a discourse (a$'$). Representations are updated when the discourse describes changes to entities' state (b$'$).}
    \label{fig:teaser}
    \vspace{-1em}
\end{figure}

Consider the text in the left column of \cref{fig:teaser}. Sentences (a) describe the contents of a room; this situation can be formally characterized by the graph of entities, properties, and relations depicted in (a$'$). Sentence (b), \emph{You pick up the key},
causes the situation to change: a chest becomes \texttt{empty}, and a key
becomes \texttt{possessed} by \emph{you} rather than \texttt{contained} by the
chest (b$'$). None of these changes are explicitly described by sentence (b).
Nevertheless, the set of sentences that can follow (a)--(b) to form a
semantically coherent discourse is determined by this new situation. An
acceptable next sentence might feature the person using the key (c$_1$) or
performing an unrelated action (c$_2$). But a sentence in which the person takes
an apple out of the chest (c$_3$) cannot follow (a)--(b), as the chest is now empty. %

Formal models of situations (built, like (a$'$)--(b$'$), from logical representations of entities and their attributes) are central to linguistic theories of meaning.
NLMs face the problem of learning to generate coherent text like (a--c) without
access to any explicit supervision for the underlying world state (a$'$)--(b$'$).
Indeed, recent work in NLP points to the lack of exposure of explicit
representations of the world external to language as \emph{prima facie} evidence
that LMs cannot represent meaning at all, and thus cannot in general output
coherent discourses like (a)--(c) \cite{bender-koller-2020-climbing}.

The present paper can be viewed as an empirical response to these arguments. 
It is true that current NLMs do not reliably output coherent descriptions when
trained on data like (a)--(c).
But from text alone, even these imperfect NLMs appear to learn \emph{implicit}
models of meaning that are translatable into formal state representations like
(a$'$)--(b$'$).
These state representations capture information like the \texttt{emptiness} of
the \texttt{chest} in (b$'$), which is not explicitly mentioned and cannot be
derived from any purely syntactic representation of (a)--(b), but follows as a
semantically necessary consequence. %
These implicit semantic models are roughly analogous to the simplest components of discourse representation theory and related formalisms: they represent sets of entities, and update the facts that are known about these entities as sentences are added to a discourse.
Like the NLMs that produce them, these implicit models are approximate and error-prone. Nonetheless, 
they do most of the things we expect of world models in formal semantics: they are structured, queryable and manipulable. %
In this narrow sense, NLM training appears to induce not just models of linguistic form, but 
models of meaning.

This paper begins with a review of existing approaches to NLM probing and
discourse representation that serve as a foundation for our approach. We then
formalize a procedure for determining whether NLM representations encode
representations of situations like \cref{fig:teaser} (a$'$)--(b$'$).
Finally, we apply this approach to BART and T5 NLMs trained on text from the English-language Alchemy and TextWorld datasets. 
In all cases, %
we find evidence of implicit meaning representations that:
\begin{enumerate}
    \item Can be linearly decoded from NLM encodings of entity mentions. %
    \item Are primarily attributable to open-domain pretraining rather than in-domain fine-tuning. 
    
    \item Influence downstream language generation.
\end{enumerate}
We conclude with a discussion of the implications of these results for evaluating and improving factuality and coherence in NLMs.

\section{Background}

\paragraph{What do LM representations encode?}

This paper's investigation of state representations builds on a large body of past work aimed at understanding 
how
other linguistic phenomena 
are represented
in large-scale language models. NLM representations have been found
to encode syntactic categories, dependency relations, and coreference information~\citep{tenney-etal-2019-bert,hewitt-manning-2019-structural,clark-etal-2019-bert}. 
Within the realm of semantics, existing work has identified representations of word meaning (e.g., fine-grained word senses; \citealt{wiedemann2019does})
and predicate--argument structures like frames and semantic roles~\citep{kovaleva-etal-2019-revealing}.

In all these studies, the main experimental paradigm is
\emph{probing}~\citep{shi-etal-2016-string,belinkov-glass-2019-analysis}: given
a fixed source of representations (e.g.\ the BERT language model;
\citealt{devlin-etal-2019-bert}) and a linguistic label of interest (e.g.\ semantic role), a low-capacity ``probe'' (e.g\ a linear classifier) is trained to predict the label from the representations (e.g.\ to predict semantic roles from BERT embeddings). A phenomenon is judged to be encoded by a model if the probe's accuracy cannot be explained by its accuracy when trained on control tasks~\cite{hewitt-liang-2019-designing} or baseline models~\cite{pimentel-etal-2020-pareto}.

Our work extends this experimental paradigm to a new class of semantic phenomena. 
As in past work, we train probes to recover semantic annotations, and interpret these probes by comparison to null hypotheses that test the role of the model and the difficulty of the task.
The key distinction is that we aim to recover a representation of the \emph{situation described by a discourse} rather than representations of the \emph{sentences that make up the discourse}. For example, in \cref{fig:teaser}, we aim to understand not only whether NLMs encode the (sentence-level) semantic information that there was a \emph{picking up} event whose patient was \emph{you} and whose agent was \emph{the key}---we also wish to understand whether LMs encode the \emph{consequences} of this action for all entities under discussion, including the chest from which the key was (implicitly) taken.

\paragraph{How might LMs encode meaning?}

\begin{figure}
    \centering
    \includegraphics[width=\columnwidth,clip,trim=0.05in 4.7in 6.85in 1.15in]{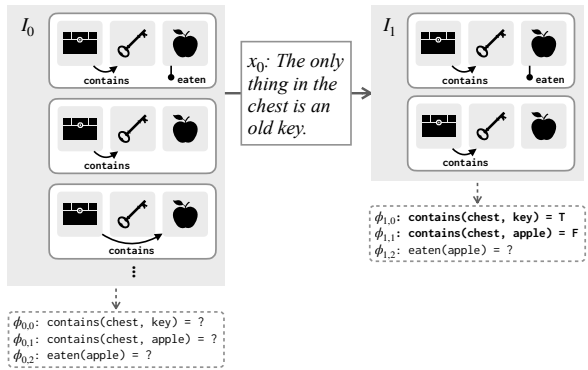}
    \vspace{-1.5em}
    \caption{A collection of possible situations is represented as an information state ($I_0$). Information states assign values to propositions $\phi_{i,j}$ according to whether they are true, false, or undetermined in all the situations that make up an information state. Appending a new sentence discourse causes the information state to be updated ($I_1$). In this case, the sentence \emph{The only thing in the chest is an old key} causes \texttt{contains(chest, key)} to become true, \texttt{contains(chest, apple)} to become false, and leaves \texttt{eaten(apple)} undetermined.}%
    \label{fig:istate}
    \vspace{-1em}
\end{figure}

Like in other probing work, an attempt to identify neural encodings of entities and situations must begin with a formal framework for representing them. This is the subject of \textbf{dynamic semantics} in linguistics~\citep{heim-2008-file-change-semantics,kamp-2010-drt,groenendijk-stokhof-dpl}.
The central tool for representing meaning in these approaches is the
\emph{information state}: the set of possible states of the world consistent with a discourse ($I_0$ and $I_1$ in \cref{fig:istate}). Before anything is said, all logically consistent situations are part of the information state ($I_0$). Each new sentence in a discourse provides an update (that constrains or otherwise manipulates the set of possible situations). As shown in the figure, these updates can affect even unmentioned entities: the sentence \emph{the only thing in the chest is a key} ensures that the proposition \texttt{contains(chest, $x$)} is false for all entities $x$ other than the key. This is formalized in \cref{sec:approach} below.\footnote{See also \citet{Yalcin2014IntroductoryNO} for an introductory survey.}

The main hypothesis explored in this paper is that \emph{LMs represent (a
particular class of) information states}. Given an LM trained on text alone, and a discourse annotated post-hoc with information states,
our probes will try to recover these information states from LM representations.
The semantics literature includes a variety of proposals for how information
states should be represented; here, we will represent information states
logically, and decode information states
via the truth values that they assign to logical propositions
($\phi_{i,j}$ in \cref{fig:istate}).\footnote{In existing work, one of the main
applications of dynamic semantics is a precise treatment of quantification and
scope at the discourse level. The tasks investigated in this paper do not
involve any interesting quantification, and rely on the simplest parts of the
formalism. More detailed exploration of quantification in NLMs is an important
topic for future study.}

\paragraph{LMs and other semantic phenomena}

    In addition to work on interpretability, a great deal of past research uses language modeling as a pretraining scheme for more conventional (supervised) semantics tasks in NLP. 
LM pretraining is useful for semantic parsing~\cite{einolghozati2019improving}, instruction following~\citep{hill2020human}, and even image retrieval~\cite{ilharco2020probing}.
Here, our primary objective is not good performance on downstream tasks, but rather understanding of representations themselves.
LM pretraining has also been found to be useful for tasks like factoid question answering~\cite{petroni-etal-2019-language,roberts-etal-2020-much}. Our experiments do not explore the extent to which LMs encode static background knowledge, but instead the extent to which they can build representations of novel situations described by novel text.

\section{Approach}

\begin{figure*}[t!]
    \centering
    \includegraphics[width=\textwidth, clip, trim=0 3.65in 1.56in 1.1in]{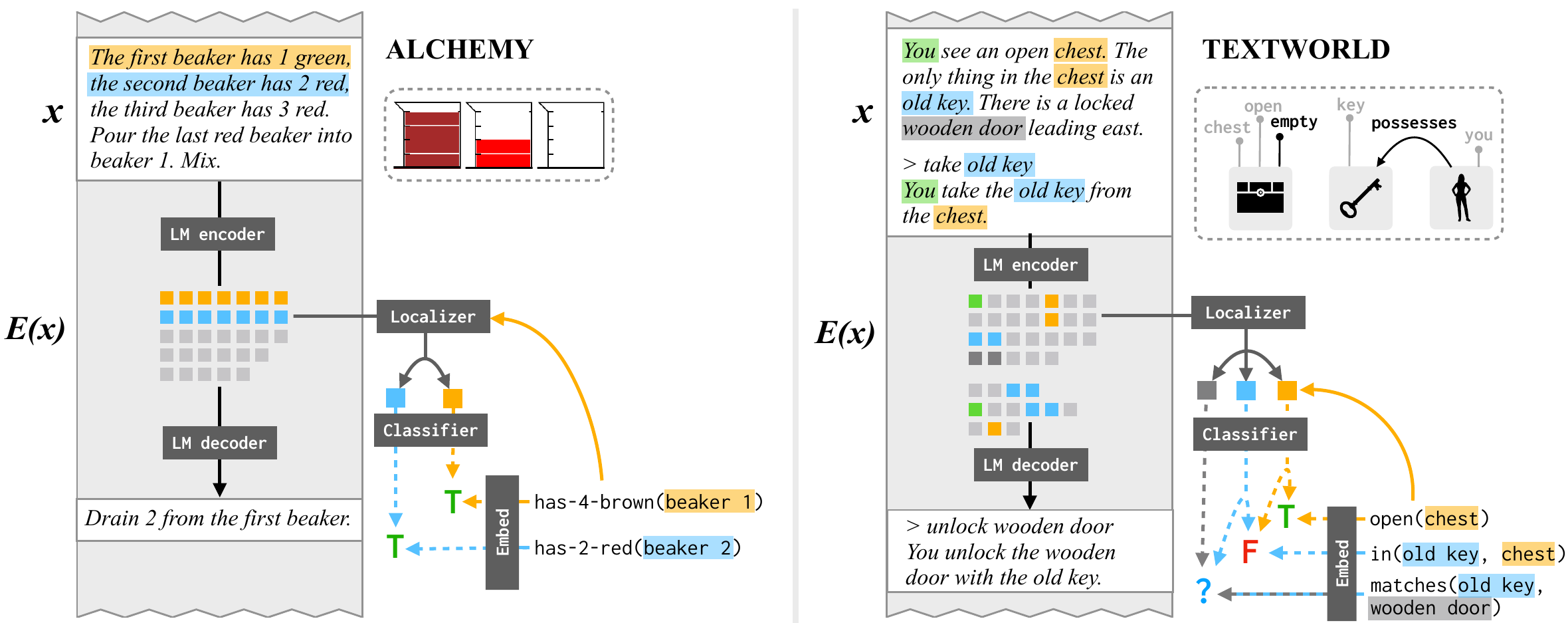}
    \caption{Overview of the probe model. \textbf{Left}: Alchemy. \textbf{Right}: Textworld. The LM is first trained to generate the next instruction from prior context (left side, both figures). Next, the LM encoder is frozen and a probe is trained to recover (the truthfulness of) propositions about the current state from specific tokens of encoder outputs.}
    \label{fig:probe}
    \vspace{-1em}
\end{figure*}

\label{sec:approach}

\paragraph{Overview} We train probing models to test
whether NLMs represent the information states specified by the input text. 
We specifically probe for the truth values of logical propositions about entities mentioned in the text. 
For example, in Figure~\ref{fig:teaser}, we test whether a representation of
sentences (a)--(b) encodes the fact that \texttt{empty(chest)} is true and \texttt{contains(chest, key)} is false. %

\paragraph{Meanings as information states} %
To formalize this: given a universe consisting of a set of entities, properties,
and relations, we define a \textbf{situation} as a \emph{complete} specification
of the properties and relations of each entity. For example, the box labeled
$I_0$ in \cref{fig:istate} shows three situations involving a \texttt{chest}, a
\texttt{key}, an \texttt{apple}, an \texttt{eaten} property and a \texttt{contains} relation. In one situation, the chest contains the key and the apple is eaten. In another, the chest contains the apple, and the apple is not eaten.
In general, a situation assigns a value of true or false to \emph{every} logical proposition of the form $P(x)$ or $R(x, y)$ (e.g.\ \texttt{locked}(\texttt{door}) or \texttt{contains}(\texttt{chest}, \texttt{key})).

Now, given a natural language discourse, we can view that discourse as specifying a \emph{set} of possible situations. In \cref{fig:istate}, the sentence $x_0$ picks out the subset of situations in which the chest contains the key.
A collection of situations is called an \textbf{information state}, because it
encodes a listener's knowledge of (and uncertainty about) the state of the world
resulting from the events described in a discourse.\footnote{An individual sentence is associated with a \emph{context change potential}: a map from information states to information states.}
In a given information state, the value of a proposition might be \textit{true} in all situations, \textit{false} in all situations, or \textit{unknown}: true in some but false in others.
An information state (or an NLM representation) can thus be characterized by the label it assigns to every proposition.%

\paragraph{Probing for propositions} 
We assume we are given:
\begin{itemize}
    \item  A sequence of sentences $x_{1:n} = [x_1, \ldots, x_n]$.
    \item For each $i$, the information state $I_i$ that results from the sentences $x_{1:i}$. We write $I(\phi) \in \{T, F, ?\}$ for the value of the proposition $\phi$ in the information state $I$. 
    \item A language model encoder $E$ that maps sentences to sequences of $d$-dimensional word representations.
\end{itemize}
To characterize the encoding of semantic information in $E(x)$, we design a semantic \textbf{probe} that tries to recover the contents of $I_i$ from $E(x_{1:i})$ proposition-by-proposition. Intuitively, this probe aims to answer three questions: (1) \textbf{How} is the truth value of a given proposition $\phi$ encoded? (Linearly? Nonlinearly? In what feature basis?) (2) \textbf{Where} is information about $\phi$ encoded? (Distributed across all token embeddings? Local to particular tokens?) (3) \textbf{How well} is semantic information encoded? (Can it be recovered better than chance? Perfectly?)

The probe is built from three components, each of which corresponds to one of the questions above:
\begin{enumerate}
    \item %
    A \textbf{proposition embedder} $\mathtt{embed} : L \to \reals^d$ (where $L$ is the set of logical propositions).
    \item A \textbf{localizer} $\mathtt{loc} : L \times \reals^d \to \reals^d$
    which extracts and aggregates LM representations 
    as candidates for encoding $\phi$. The localizer extracts tokens of $E(x)$ at positions corresponding to particular tokens in the underlying text $x$. We express this in notation as $E(x)[\textit{*}]$, where \textit{*} is a subsequence of $x$. 
    (For example, if $x =$ ``\textit{the third beaker is empty}". $E(x) = [v_1,v_2,v_3,v_4,v_5]$ has one vector per token. $E(x)[``\textit{third beaker}"] = [v_2,v_3]$.)
    \item A \textbf{classifier} $\mathtt{cls}_\theta : \reals^d \times \reals^d \to \{T, F, ?\}$, which takes an embedded proposition and a localized embedding, and predicts the truth value of the proposition.
\end{enumerate}
\noindent
We say that a proposition $\phi$ is encoded by $E(x)$ if:
\begin{align}
    \label{eq:probe}
    \begin{split}
    \mathtt{cls}_\theta(\mathtt{embed}(\phi), \mathtt{loc}(\phi,E(x))) = I(\phi) ~ .
    \end{split}
\end{align}
Given a dataset of discourses $\mathcal D$, we 
attempt to find a classifier parameters $\theta$ from which all propositions can
be recovered for all sentences in \cref{eq:probe}. To do so, we
label each with the
truth/falsehood of every relevant proposition. We then train the parameters of a
$\mathtt{cls}_\theta$ on a subset of these propositions and test whether it
generalizes to held-out discourses.

\section{Experiments}
\label{sec:experiments}

Our experiments aim to discover to what extent (and in what manner) information
states are encoded in NLM representations. %
We first present a specific instantiation of the probe that allows us to
determine how well information states are encoded in two NLMs and two datasets
(\cref{sec:probe});
then provide a more detailed look at \emph{where} specific propositions are encoded by varying
$\texttt{loc}$ (\cref{sec:localization}).
Finally, we describe an experiment investigating the causal role of semantic
representations by directly manipulating %
$E(x)$ (\cref{sec:alchemy_intervene}).\footnote{Sections here are also discussed in more detail in Appendix~\ref{sec:appendix_data_details} (for \S\ref{sec:preliminaries}), \ref{sec:appendix_probe_details} (for \S\ref{sec:probe}), and \ref{sec:appendix_localization} (for \S\ref{sec:localization}).}

\subsection{Preliminaries}
\label{sec:preliminaries}
\paragraph{Model}
In all experiments, the encoder $E$ comes from a BART~\cite{lewis-etal-2020-bart}
or T5~\cite{2020t5} model.
Except where noted, 
BART is pretrained on OpenWebText, BookCorpus, CC-News, and Stories~\cite{lewis-etal-2020-bart}, 
T5 is pretrained on C4~\cite{2020t5},
and both are fine-tuned on the TextWorld or Alchemy datasets described below. Weights of $E$ are frozen %
during probe training.

\paragraph{Data: Alchemy}
Alchemy, the first dataset used in our experiments, is derived from the SCONE~\cite{long-etal-2016-simpler} semantic parsing tasks. We preserve the train / development split from the original dataset (3657 train / 245 development). Every example in the dataset consists of a human-generated sequence of instructions to drain, pour, or mix a beaker full of colored liquid. Each instruction is annotated with the ground-truth state that results from following that instruction (\autoref{fig:probe}).

We turn Alchemy into a language modeling dataset by prepending a declaration of the initial state (the initial contents of each beaker) to the actions. 
The initial state declaration 
always follows a fixed form (``\textit{the first beaker has \emph{[amount]
[color]}, the second beaker has \emph{[amount] [color]}, ...}"). Including it in the context provides enough information that it is (in principle)
possible to deterministically compute the contents of each beaker after each instruction. 
The NLM is trained to predict the next instruction based on a textual description of the initial state and previous instructions.

The state representations we probe for in Alchemy describe the contents of each beaker. Because execution is deterministic and the initial state is fully specified, the information state associated with each instruction prefix consists of only a single possible situation, 
 defined by 
a set of propositions:
\begin{align}
    \label{eq:alchemy_facts}
    \begin{split}
    \Phi = &\big\{ \texttt{has-$v$-$c$}(b) : \\
    & b \in \{\texttt{beaker 1}, \texttt{beaker 2}, \ldots\}, \\
    & v \in 1..4, \\
    & c \in \{\texttt{red}, \texttt{orange}, \texttt{yellow}, \ldots\}\big\} ~ .
    \end{split}
\end{align}
In the experiments below, it will be useful to have access to a natural language
representation of each proposition. We denote this:
\begin{align}
    \label{eq:alchemy_facts_nl}
    \texttt{NL}(\texttt{has-$v$-$c$}(b)) = ``\textit{the $b$ beaker has $v$ $c$}"
    ~ .
\end{align}
Truth values for each proposition in each instruction sequence are %
straightforwardly derived from ground-truth state annotations in the dataset.

\paragraph{Data: TextWorld}
TextWorld~\cite{cote18textworld} is a platform for generating synthetic worlds for text-based games, used to test RL agents. The game generator produces rooms containing objects, surfaces, and containers, which the agent can interact with in various predefined ways.%

\begin{table*}[]
    \centering
    \footnotesize
    \begin{tabular}{crcccccccc}
    \toprule
        & & \multicolumn{4}{c}{\textbf{Alchemy}} & \multicolumn{4}{c}{\textbf{TextWorld}} \\
    \cmidrule(lr){3-6}\cmidrule(lr){7-10}
        & & \multicolumn{2}{c}{State EM} & \multicolumn{2}{c}{Entity EM} & 
        \multicolumn{2}{c}{State EM} & \multicolumn{2}{c}{Entity EM} \\
    \cmidrule(lr){3-4}\cmidrule(lr){5-6}\cmidrule(lr){7-8}\cmidrule(lr){9-10}
        & & BART & T5 & BART & T5 & BART & T5 & BART & T5 \\
    \midrule
        main probe (\S\ref{sec:probe}) &
        & 7.6 & 14.3 & 75.0 & 75.5 & %
        48.7 & 53.8 & 95.2 & 96.9 \\
    \midrule
        & +pretrain, -fine-tune & 1.1 & ~~4.3 & 69.3 & 74.1 & %
        23.2 & 38.9 & 91.1 & 94.3 \\
        baselines \& & -pretrain, +fine-tune & \multicolumn{2}{c}{1.5} & \multicolumn{2}{c}{62.8} & %
        \multicolumn{2}{c}{14.4} & \multicolumn{2}{c}{81.2} \\
        model ablations & random init. & \multicolumn{2}{c}{0.4} & \multicolumn{2}{c}{64.9} & %
        \multicolumn{2}{c}{11.3} & \multicolumn{2}{c}{74.5} \\
        (\S\ref{sec:probe}) & no change & \multicolumn{2}{c}{0.0} & \multicolumn{2}{c}{62.7} & \multicolumn{2}{c}{9.73} & \multicolumn{2}{c}{74.1} \\
        & no LM & \multicolumn{2}{c}{0.0} & \multicolumn{2}{c}{32.4} & %
        \multicolumn{2}{c}{1.77} & \multicolumn{2}{c}{81.8} \\
    \midrule
        \multirow{2}{*}{locality (\S\ref{sec:localization})} 
        & first & \multicolumn{2}{c}{-} & \multicolumn{2}{c}{-} & 49.6 & 51.5 & 93.6 & 95.9 \\
        & last & \multicolumn{2}{c}{-} & \multicolumn{2}{c}{-} & 55.1 & 58.6 & 96.5 & 97.6 \\
    \bottomrule
    \end{tabular}
    \vspace{-.3em}
    \caption{Probing results. For each dataset, we report \textit{Entity EM},
    the \% of entities for which all propositions were correct, and
    \textit{State EM}, the \% of states for which all proposition were correct.
    For non-pretrained baselines (\emph{-pretrain, +fine-tune} and \emph{random
    init.}), we report the single best result from all model configurations
    examined.
    Semantic state information can be recovered at the entity level from both
    language models on both datasets, and successful state modeling appears to
    be primarily attributable to pretraining rather than fine-tuning.
    } %
    \label{tab:probe_main}
    \vspace{-1em}
\end{table*}

We turn TextWorld into a language modeling dataset by generating random game rollouts following the ``simple game" challenge, which samples world states with a fixed room layout but changing object configurations. 
For training, we sample 4000 rollouts across 79 worlds, and for development, we sample 500 rollouts across 9 worlds.
Contexts begin with a description of the room that the player currently stands in, and all visible objects in that room. This is followed by a series of actions (preceded by \verb|>|) and game responses (\cref{fig:probe}).

The NLM is trained
to generate both an action and a game response from a history of interactions. 

We probe for both the properties of and relations between entities at the end of a sequence of actions.
Unlike Alchemy, these may be undetermined, %
as the agent may not have explored the entire environment by the end of an action sequence.
(For example, in \cref{fig:probe}, the truth value of \texttt{matches(old key, door)} is \textit{unknown}).
The set of propositions available in the TextWorld domain has form
\begin{align}
    \label{eq:tw_facts}
    \begin{split}
    \hspace{-1em} \Phi = &\{ p(o) : o \in O, p \in P \} \\
    & \cup \{ r(o_1, o_2) : o_1, o_2 \in O, r \in R \}
    \end{split}
\end{align}
for objects $O = \{\texttt{player}, \texttt{chest}, \ldots\}$, properties $P = \{\texttt{open}, \texttt{edible}, \ldots\}$ and relations $R = \{\texttt{on}, \texttt{in}, \ldots\}$. 
We convert propositions to natural language descriptions as:
\begin{align}
    \label{eq:tw_facts_nl}
    \begin{split}
    \texttt{NL}(p(o)) & = ``\textit{the $p$ is $o$}" \\
    \texttt{NL}(r(o_1, o_2)) & = ``\textit{the $o_1$ is $r$ $o_2$}" ~ .
    \end{split}
\end{align}
The set of propositions and their natural language descriptions are pre-defined by TextWorld's simulation engine.
The simulation engine also gives us the set of true propositions, from which we can 
compute the set of false and unknown propositions.

\paragraph{Evaluation}
We evaluate probes according to two metrics. %
\textbf{Entity Exact-Match (EM)} first aggregates the propositions by \textit{entity} or \emph{entity pair}, then counts the percentage of entities for which \textit{all} propositions were correctly labeled. \textbf{State EM} aggregates propositions by \textit{information state} (i.e.\ context), then counts the percentage of states for which all facts were correctly labeled.

\subsection{Representations encode entities' final properties and relations}
\label{sec:probe}

With this setup in place, we are ready to ask our first question: is semantic state information encoded at all by 
pretrained LMs fine-tuned on Alchemy and TextWorld?
We instantiate the probing experiment defined in \cref{sec:approach} as follows:

    The \textbf{proposition embedder} 
    converts each proposition $\phi\in \Phi$ to its natural language description,
    embeds it using the same LM encoder that is being probed, then averages the
    tokens:
    \begin{equation}\label{eq:embed}\texttt{embed}(\phi) =
    \texttt{mean}(E(\texttt{NL}(\phi)))\end{equation}
    
    The \textbf{localizer} associates each proposition $\phi$ with %
    specific tokens corresponding to  the entity or entities that $\phi$ describes, then averages these tokens. 
    In Alchemy, we average over tokens in the \emph{initial description} of the beaker in question.
    For example, let $x$ be the discourse in Figure~\ref{fig:probe} (left) and $\phi$ be a proposition about the first beaker.
    Then, e.g., %
    \begin{equation}
    \begin{split}
        &\texttt{loc}(\texttt{has-1-red(beaker 1)}, E(x)) = \\
        &~~\texttt{mean}(E(x)[\textit{The first beaker has 2 green,}]).
        \end{split}
    \end{equation}
    In TextWorld, we average over tokens in \emph{all mentions} of each entity. Letting $x$ be the discourse in Figure~\ref{fig:probe} (right), we have:
    \begin{equation}
        \begin{split}
            &\texttt{loc}(\texttt{locked(wooden door)}, E(x)) = \\
            &\quad \texttt{mean}(E(x)[\textit{wooden door}]) ~ .
        \end{split}
    \end{equation}
    Relations, with two arguments, are localized by taking the mean of the two mentions.

    Finally, the \textbf{classifier} is a \emph{linear model} which maps each NLM representation and proposition to a truth value.
    In Alchemy, a linear transformation is applied to the NLM representation, and then the proposition with the maximum dot product with that vector is labelled $T$ (the rest are labelled $F$).
    In TextWorld, a bilinear transformation maps each (proposition embedding, NLM representation) pair to a distribution over $\{T, F, ?\}$.
    
    As noted by \citet{liang2015bringing}, it is easy to construct examples of
    semantic judgments that cannot be expressed as linear functions of purely
    syntactic sentence representations. We expect (and verify with ablation
    experiments) that this probe is not expressive enough to compute
    information states directly from surface forms, and \emph{only} expressive
    enough to read out state information already computed by the underlying LM.

\begin{figure}
    \centering
    \label{fig:locality}
    \includegraphics[width=\columnwidth, clip, trim=0 12.5in 21.1in 0]{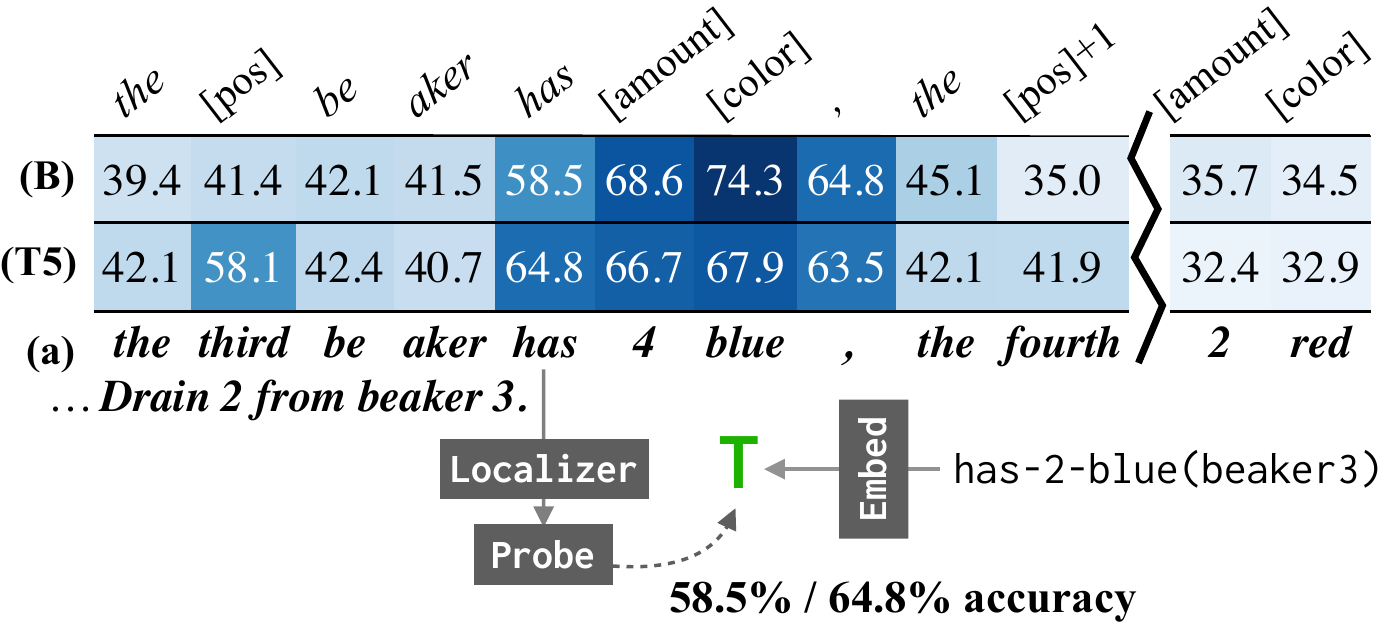}
    \caption{
    Locality of information state in Alchemy.
    We focus on the initial state declaration.
    Linear probes are trained to decode the final state of a beaker conditioned on the individual contextualized representations for each word. %
    Separate probes are trained for each position.
    Tokens in the same relative position (with respect to the target beaker) are superimposed and the averaged entity EM is reported in (B) for BART and (T5) for T5. (a) shows the paradigm on a concrete example. 
    }
    \vspace{-1em}
    \label{fig:alchemy_tokens}
\end{figure}

\paragraph{Results}
Results are shown in \cref{tab:probe_main}. 
A probe on T5 can exactly recover 14.3\% of information states in Alchemy, and 53.8\% in TextWorld.
For context, we compare to two \textbf{baselines}: a \emph{no LM} baseline, which simply predicts the most frequent final state for each entity, and a \emph{no change} baseline, which predicts that the entity's final state in the discourse will be the same as its initial state.
The \emph{no LM} baseline is correct 0\% / 1.8\% of the time and the \emph{no change} baseline is %
correct 0\% / 9.7\% of the time---substantially lower than the main probe.

To verify that this predictability is a property of the NLM representations rather than the text itself, 
we apply our probe to a series of \textbf{model ablations}. First, we evaluate a
\emph{randomly initialized} transformer %
rather than the pretrained and fine-tuned model, which has much lower probe accuracy.
To determine whether the advantage is conferred by LM pretraining or fine-tuning,
we ablate either open-domain pretraining, in a
\emph{\hbox{-pretrain},\hbox{+fine-tune}}
ablation, or in-domain fine-tuning, in a \emph{\hbox{+pretrain},\hbox{-fine-tune}} ablation. %
(For all experiments \emph{not} using a pretrained model checkpoint, we
experimented with both a BART-like and T5-like choice of depth and hidden size,
and found that the BART-like model performed better.)
While both fine-tuning and pretraining contribute to the final probe accuracy,
pretraining appears to play a much larger role: {semantic state can be
recovered well from models with no in-domain fine-tuning}.

Finally, we note that there may be lexical overlap between the discourse and
natural language descriptions of propositions. How much of the probe's
performance can be attributed to this overlap? In Alchemy, the \emph{no change}
baseline (which performs much worse than our probe) also acts as a lexical
overlap baseline---there will be lexical overlap between true propositions and
the initial state declaration only if the beaker state is unchanged. In
TextWorld, each action induces multiple updates, but can at most overlap with
one of its affected propositions (e.g. \emph{You close the chest} causes
\texttt{closed(chest)} and $\neg$\texttt{open(chest)}, but only overlaps with
the former). Moreover, only $\sim$50\% of actions have lexical
overlap with any propositions at all. Thus, lexical overlap cannot fully explain probe performance in either domain.

In summary, \emph{pretrained NLM representations model state changes and encode
semantic information about entities' final states}.

\subsection{Representations of entities are local to entity mentions}
\label{sec:localization}
\begin{table}[]
    \centering
    \small
    \begin{tabular}{lcccc}
    \toprule
        & \multicolumn{2}{c}{State EM} & \multicolumn{2}{c}{Entity EM} \\
    \cmidrule(lr){2-3}\cmidrule(lr){4-5}
        & BART & T5 & BART & T5 \\
    \midrule
        remap & 50.2 & 50.4 & 88.9 & 93.2 \\
        main probe & 50.2 & 53.8 & 91.3 & 94.6 \\
    \bottomrule
    \end{tabular}
    \vspace{-.3em}
    \caption{Locality of information state in TextWorld (T5). 
    Entity state information tends to be slightly more present in mentions of the target entity (main probe) rather than of other entities (remap), but not by much. %
    }
    \label{tab:tw_remap}
    \vspace{-1em}
\end{table}

The experiment in \cref{sec:probe} assumed that entity state could be recovered from a fixed set of input tokens. Next, we conduct a more detailed investigation into \textit{where} state information is localized.
To this end, we ask two questions: first, can we assume state information is localized in the corresponding entity mentions, and second, if so, \textit{which} mention encodes the most information, and what kind of information does it encode?

\subsubsection{Mentions or other tokens?}
\label{sec:which_token}
We first contrast tokens \emph{within} mentions of the target entity to tokens \emph{elsewhere} in the input discourse.
In Alchemy, each beaker $b$'s initial state declaration is tokenized as:
$\texttt{toks}_b$ = \{\emph{the$_b$, \emph{[position]}$_b$, be$_b$, aker$_b$, has$_b$, \emph{[volume]}$_b$, \emph{[color]}$_b$, ,$_b$}\}, where $b$ signifies the beaker position.
Rather than pooling these tokens together (as in \cref{sec:probe}), we 
construct a \textbf{localizer ablation} that
associates beaker $b$'s state with \emph{single tokens} $t$ in either the initial mention of beaker $b$, or the initial mention of \emph{other beakers} at an integer offset $\Delta$.
For each $(t,\Delta)$ pair, we construct a localizer that matches propositions about beaker $b$ with $t_{b+\Delta}$. For example, the $(\textit{has}, +1)$ localizer associates the \emph{third} beaker's final state with the vector in $E(x)$ at the position of the ``has" token in \textit{the fourth beaker has 2 red}.

In TextWorld, which does not have such easily categorizable tokens, we investigate whether information about the state of an entity is encoded in mentions of \textit{different entities}. 
We sample a random mapping $\texttt{remap}$ between entities, and construct a localizer ablation in which we decode propositions about $w$ from mentions of $\texttt{remap}(w)$.
For example, we probe the value of \texttt{open(chest)} from mentions of \textit{old key}.
These experiments use a different evaluation set%
---we restrict evaluation to the subset of entities for which \textit{both} $w$ and \texttt{remap}$(w)$ appear in the discourse. For comparability, we re-run the main probe on this restricted set.\footnote{The \texttt{remap} and $\Delta \neq 0$ probes described here are analogous to \textit{control tasks} \citep{hewitt-liang-2019-designing}. They measure probes' abilities to predict labels that are structurally similar but semantically unrelated to the phenomenon of interest.}

\paragraph{Results}
Fig.~\ref{fig:alchemy_tokens} shows the locality of BART and T5 in the Alchemy domain. 
Entity EM is highest for %
words corresponding to the correct beaker, and specifically for color words. %
Decoding from any token of an incorrect beaker barely outperforms the \textit{no LM} baseline (32.4\% entity EM).
In TextWorld, Table~\ref{tab:tw_remap} shows that decoding from a remapped entity is only 1-3\% worse than decoding from the right one.
Thus, \emph{the state of an entity $e$ is (roughly) localized to tokens in
mentions of $e$}, though the degree of locality is data- and model-dependent.

\subsubsection{Which mention?}
\label{sec:which_mention}
To investigate facts encoded in different mentions of the entity in question,
we experiment with decoding from the first and last mentions of the entities in $x$.
The form of the localizer is the same as~\ref{sec:probe}, except
instead of averaging across all mentions of entities, we use the first mention 
or the last mention.
We also ask whether \textit{relational} 
propositions can be decoded from just one argument (e.g., \texttt{in(old key, chest)} from just mentions of \textit{old key}, rather than the averaged encodings of \textit{old key} and \textit{chest}).

\paragraph{Results}
As shown in Table~\ref{tab:probe_main}, 
in TextWorld, probing the \textit{last} mention gives the highest accuracy. Furthermore, as Table~\ref{tab:textworld_simple_relations}  shows, \emph{relational facts can be decoded from either side of the relation}. %

\begin{table}[t!]
    \small
    \centering
    \begin{tabular}{rcccc}
    \toprule
        & \multicolumn{2}{c}{Relations} & \multicolumn{2}{c}{Properties} \\
    \cmidrule(lr){2-3}\cmidrule(lr){4-5}%
        & BART & T5 & BART & T5 \\ %
    \midrule
        1-arg & 41.4 & 55.5 & 83.2 & 90.9 \\
        2-arg & 49.6 & 54.4 & 94.5 & 98.5 \\
        random init. & \multicolumn{2}{c}{21.9} & \multicolumn{2}{c}{35.4} \\
    \bottomrule
    \end{tabular}
    \vspace{-.3em}
    \caption{State EM for decoding each type of fact (relations vs. properties), with each type of probe (1- vs. 2- argument decoding). Though decoding from two entities is broadly better, one entity still contains a non-trivial amount of information, even regarding relations.}
    \vspace{-1em}
    \label{tab:textworld_simple_relations}
\end{table}

\subsection{Changes to entity representations cause changes in language model predictions}
\label{sec:alchemy_intervene}

The localization experiments in Section~\ref{sec:localization} indicate that state information is localized within contextual representations in predictable ways. This suggests that modifying the %
representations themselves could induce systematic and predictable changes in model behavior.
We %
conduct a series of causal intervention experiments in the Alchemy domain which
measure effect of \emph{manipulating encoder representations} on NLM output.
We \textit{replace} a small subset of token representations with those from a different information state, and show that this causes the model to behave as if it were in the new information state.\footnote{This experiment is inspired by~\citet{geiger-etal-2020-neural}.}

\begin{figure}
    \centering
    \includegraphics[width=\columnwidth, clip, trim=0 2.73in 6.4in 1.1in]{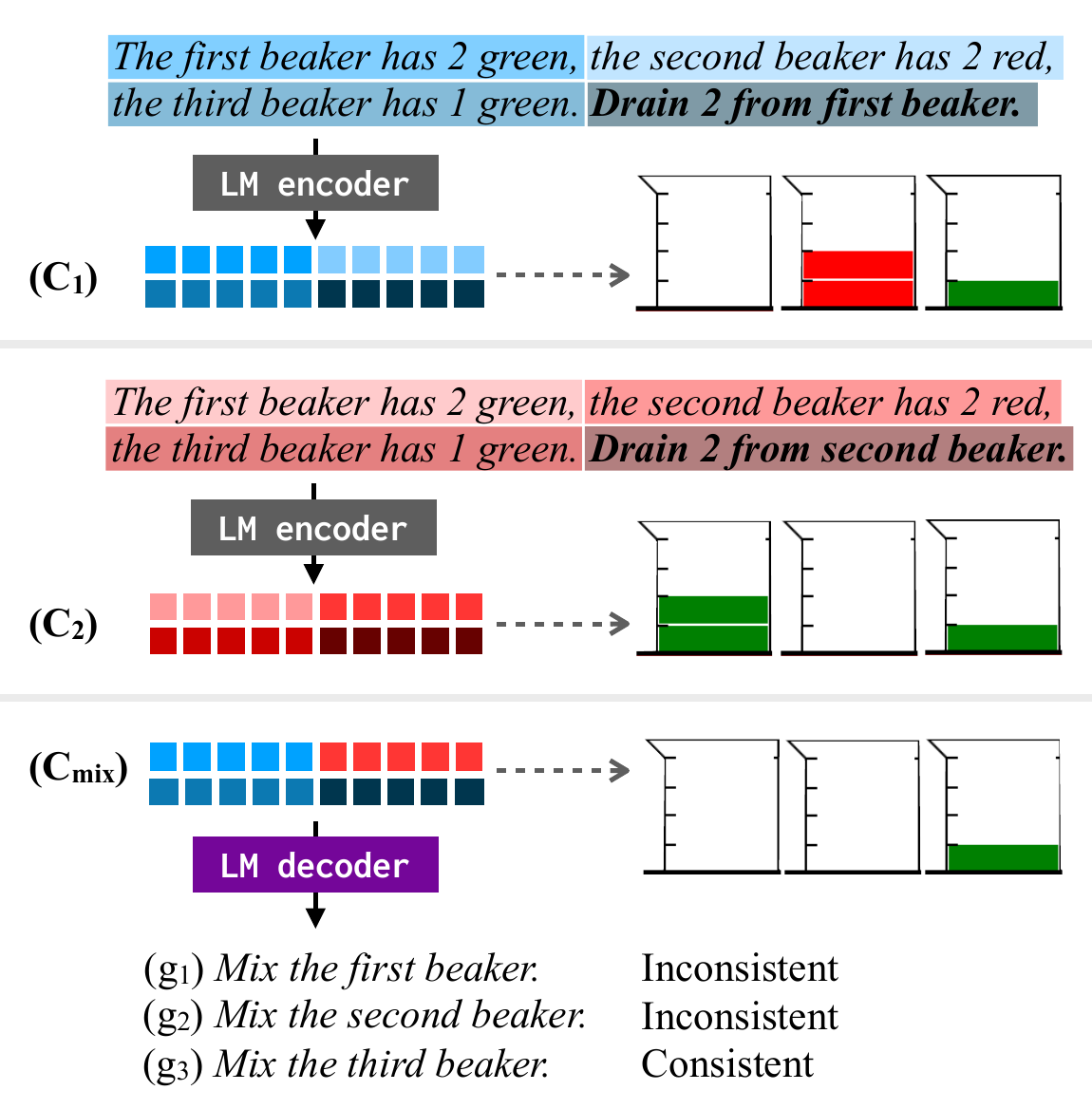}
    \vspace{-2em}
    \caption{Intervention experiments. Construct $C_1,C_2$ by appending text to empty one of the beakers (e.g. the first and second beakers) 
    and encoding the result.
    Then, create $C_\text{mix}$ by taking encoded tokens from $C_1$ and replacing the encodings corresponding to the second beaker's initial state declaration with those from $C_2$. This induces the LM to model both the first and second beakers as empty, and the LM decoder should generate actions consistent with this state.}
    \label{fig:alchemy_intervene}
    \vspace{-1em}
\end{figure}

A diagram of the procedure is shown in Fig.~\ref{fig:alchemy_intervene}. 
We create two discourses, $x_1$ and $x_2$, in which \textit{one} beaker's final volume is zero.
Both discourses describe the same initial state, %
but for each $x_i$, we append the sentence \textit{drain $v_i$ from beaker $b_i$}, where $v_i$ is the initial volume of beaker $b_i$'s contents.
Though the underlying initial state tokens are the same, we expect the contextualized representation $C_1 = E(x_1)[\textit{the }i\textit{th beaker}\ldots]$ to differ from $C_2 = E(x_2)[\textit{the }i\textit{th beaker}\ldots]$ due to the different final states of the beakers. %
\newcommand{\cont}[1]{\textsc{cont}(#1)}
Let $\cont{x}$ denote the set of sentences constituting
\emph{semantically acceptable continuations} of a discourse prefix $x$. (In
\cref{fig:teaser}, $\cont{\textrm{a, b}}$ contains c$_1$ and c$_2$ but not
c$_3$.)%
\footnote{In order to automate evaluation of consistency, we use a version of Alchemy with synthetically generated text. The underlying LM has also been fine-tuned on synthetic data.} %
In Alchemy, $\cont{x_1}$ should not contain mixing, draining, or pouring
actions involving $b_1$ (similarly for $\cont{x_2}$ and $b_2$).
Decoder samples given $C_i$ should fall into $\cont{x_i}$.%

Finally, we \emph{replace the encoded description of beaker 2 in $C_1$ with its encoding from $C_2$}, creating a new representation $C_\text{mix}$. $C_\text{mix}$ was not derived from any real input text, but implicitly represents a situation in which \textit{both} $b_1$ and $b_2$ are empty.
A decoder generating from $C_\text{mix}$ should generate instructions in
$\cont{x_1} \cap \cont{x_2}$
to be consistent with this situation.

\paragraph{Results}
We generate instructions %
conditioned on $C_\text{mix}$ and check whether they are in the expected sets. 
Results, shown in Table~\ref{tab:alchemy_replace_consistency}, align with this prediction.
For both BART and T5, substantially more generations from
$C_\text{mix}$ fall within $\cont{x_1} \cap \cont{x_2}$ than from $C_1$ or
$C_2$. Though imperfect (compared to $C_1$ generations within $\cont{x_1}$ and
$C_2$ generations within $\cont{x_2}$), this
suggests that the information state associated with the synthetic encoding
$C_\text{mix}$ is (approximately) one in which \emph{both} beakers are empty.

\begin{table}[]
    \scriptsize
    \centering
    \begin{tabular}{lcccccc}
    \toprule
        & \multicolumn{6}{c}{\% of generations within...} \\
    \cmidrule(lr){2-7}
        & \multicolumn{2}{c}{$\cont{x_1} \cap \cont{x_2}$} &
        \multicolumn{2}{c}{$\cont{x_1}$} & \multicolumn{2}{c}{$\cont{x_2}$} \\
    \cmidrule(lr){2-3}\cmidrule(lr){4-5}\cmidrule(lr){6-7}
        & BART & T5 & BART & T5 & BART & T5 \\
    \midrule
        $C_1$ & 20.4 & 37.9 & \textbf{96.2} & \textbf{93.0} & 21.6 & 40.8 \\
        $C_2$ & 16.1 & 29.1 & 24.1 & 37.9 & \textbf{87.7} & \textbf{87.2} \\
        $C_\text{mix}$ & \textbf{57.7} & \textbf{75.4} & 86.7 & 86.8 & 64.8 & 84.5 \\
    \bottomrule
    \end{tabular}
    \vspace{-.3em}
    \caption{Results of intervention experiments. %
    Though imperfect, generations from $C_\text{mix}$ are more often consistent
    with both contexts compared to those from $C_1$ or $C_2$, indicating that
    its underlying information state (approximately) models both beakers as empty.
    }
    \vspace{-1em}
    \label{tab:alchemy_replace_consistency}
\end{table}

\section{Limitations}
\label{sec:limitations}
\paragraph{...of large NLMs:}
It is important to emphasize that both LM output 
and implicit state representations are imperfect: even in the best case, complete information states can only be recovered 53.8\% of the time in tasks that most humans would find very simple. (Additional experiments described in Appendix~\ref{sec:appendix_error_analysis} offer more detail about these errors.) The success of our probing experiments should not be taken to indicate that the discovered semantic representations have anything near the expressiveness needed to support human-like generation. %

\paragraph{...of our experimental paradigm:}
While our probing experiments in \cref{sec:probe} provide a detailed picture of structured state representations in NLMs, the intervention experiments in \cref{sec:alchemy_intervene} explain the relationship between these state representations and model behavior in only a very general sense. %
They leave open the key question of whether \emph{errors} in language model prediction are attributable to errors in the underlying state representation. %
Finally, the situations we model here are extremely simple, featuring just a handful of objects.  
Thought experiments on the theoretical capabilities of NLMs 
(e.g.\ \citeauthor{bender-koller-2020-climbing}'s ``coconut catapult'') involve far richer worlds and more complex interactions.
Again, we leave for future work the question of whether current models can learn to represent them.

\section{Conclusion}

Even when trained only on language data, NLMs encode simple representations of meaning. In experiments on two domains, internal representations of text produced by two pretrained language models can be mapped, using a linear probe, to representations of 
the state of the world described by the text. These internal representations are structured, interpretably localized, and editable. %
This finding has important implications for research aimed at improving factuality and and coherence in NLMs: future work might probe LMs for the the states and properties ascribed to entities the first time 
they are mentioned 
(which may reveal biases learned from training data; \citealt{bender2021dangers}), or 
correct errors in generation by directly editing representations.

\section*{Acknowledgments}

Thanks to Ekin Akyürek, Evan Hernandez, Joe O'Connor, and the anonymous reviewers for feedback on early versions of
this paper. MN is supported by a NSF Graduate Research Fellowship.
This work was supported by a hardware donation from NVIDIA under the
NVAIL program.

\section*{Impact Statement}
This paper investigates the extent to which neural language models build meaning representations of the world, and introduces a method to probe and modify the underlying information state. We expect this can be applied to improve factuality, coherence, and reduce bias and toxicity in language model generations. Moreover, deeper insight into how neural language models work and what exactly they encode can be important when deploying these models in real-world settings.

However, interpretability research is by nature dual-use and improve the effectiveness of models for generating false, misleading, or abusive language. Even when not deliberately tailored to generation of harmful language, learned semantic representations might not accurately represent the world because of errors both in prediction (as discussed in \cref{sec:limitations}) and in training data.

\bibliography{anthology,acl2021}
\bibliographystyle{acl_natbib}

\clearpage
\appendix

\section{Appendix}
\label{sec:appendix}

\subsection{Datasets Details (\S\ref{sec:preliminaries})}
\label{sec:appendix_data_details}
\paragraph{Alchemy} Alchemy is downloadable at \url{https://nlp.stanford.edu/projects/scone/}. Alchemy propositions are straightforwardly derived from existing labels in the dataset. We preserve the train/dev split from the original dataset (3657 train/245 dev), which we use for training the underlying LM \textit{and} the probe. 
In subsequent sections, we include additional results from a synthetic dataset that we generated (3600 train/500 dev), where actions are created following a fixed template, making it easy to evaluate consistency.

\paragraph{Textworld}
We generate a set of worlds for training, and a separate set of worlds for testing. We obtain transcripts from three agents playing on each game: a perfect agent and two (semi-)random agents, which intersperse perfect actions with several steps of random actions. For training, we sample 4000 sequences from the 3 agents across 79 worlds. For development, we sample 500 sequences from the 3 agents across 9 worlds.

During game generation, we are given the set of all propositions that are \textit{True} in the world, and how the set updates after each player action.
However, the player cannot infer the \textit{full} state before interacting with and seeing all objects, and neither (we suspect) can a language model trained on partial transcripts.
For example, a player that starts in the bedroom cannot infer \texttt{is-in(refrigerator, kitchen)} without first entering the kitchen.
One solution would be to hard-code rules--a player can only know about the states of entities it has directly affected or seen.
However, since synthetically-generated worlds might share some commonalities, a player that has played many games before (or an LM trained on many transcripts) might be able to draw particular conclusions about entities in unseen worlds, even before interacting with them.

To deal with these factors, we train a \textbf{labeller} model $\texttt{label}$ to help us classify propositions as \textit{known true}, \textit{known false}, and \textit{unknown}.
We generate a training set (separate from the training set for the probe and probed LM) to train the labeller.
We again use BART, but we give it the text transcripts and train it to directly decode the full set of True propositions and False proposition by the end of the transcript (recall we have the ground-truth full True set, and we label all propositions that aren't in the True set as False). 
This allows the labeller model to pick up both patterns \textit{between the discourse and its information state}, as well as infer \textit{general patterns among various discourses}.
Thus, on unknown worlds, given text $T$, if proposition $A$ is True most or all of the time given $T$, the model should be confident in predicting $A$. We label $A$ as True in these cases. However, if proposition $A$ is True only half of the time given $T$, the model is un-confident. We label $A$ as Unknown in these cases. 
Thus, we create our \textit{unknown} set using a confidence threshold $\tau$ on $\texttt{label}$'s output probability.

\subsection{Probe Details + Additional Results (\S\ref{sec:probe})}
\label{sec:appendix_probe_details}
\begin{figure*}[t!]
    \centering
    \includegraphics[width=\textwidth, clip, trim=0in 0.67in -0.2in 0.4in]{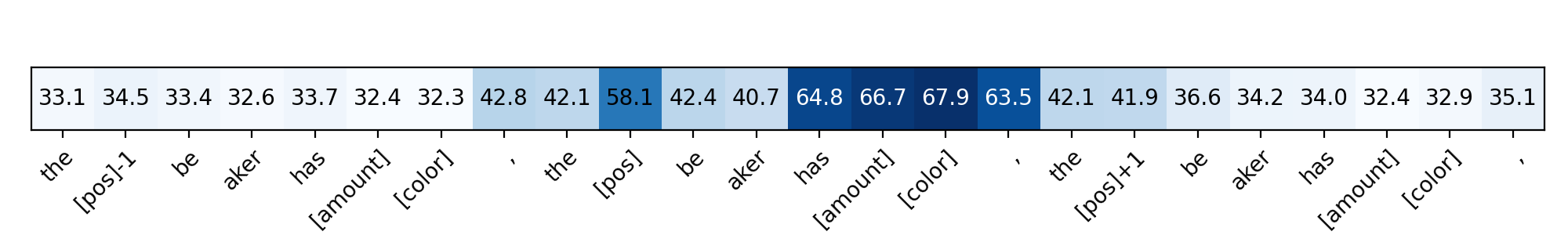}
    \includegraphics[width=\textwidth, clip, trim=0in 0.67in -0.2in 0.4in]{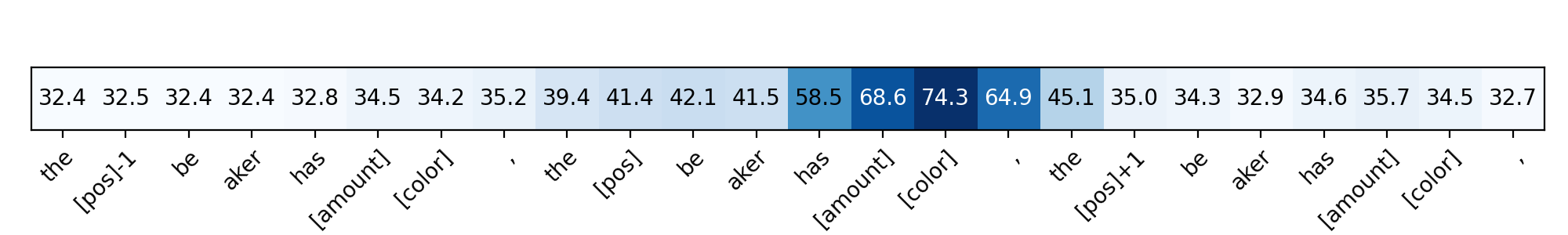}
    \includegraphics[width=\textwidth, clip, trim=0in 0in -0.2in 0.4in]{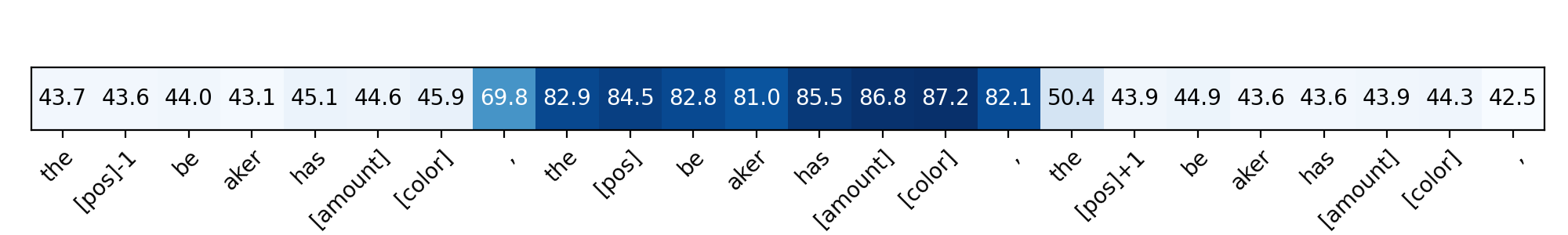}
    \caption{Alchemy locality - full results. \textbf{Top}: T5, Finetuned+probed on real data. \textbf{Middle}: BART, Finetuned+probed on real data. \textbf{Bottom}: BART, Finetuned+probed on synthetic data. We note that for the synthetic data, accurate decoding is possible from a much wider set of tokens, but all still correspond to the relevant beaker.}
    \vspace{-1em}
    \label{fig:alchemy_tokens_full}
\end{figure*}
Below, we give a more detailed account of our probing paradigm in each domain, including equations.

\paragraph{Alchemy Probe}
The proposition embedder converts propositions $\phi$ to natural language descriptions %
$\widetilde{\phi}$ (``\texttt{the $b$th beaker has $v$ $c$}") and encodes them with the BART or T5 LM encoder.

Given a proposition \texttt{has-$v$-$c$}$(b)$, 
the localizer $\texttt{loc}$ maps \texttt{has-$v$-$c$}$(b)$ to tokens in $E(x)$ that corresponding to the initial state of beaker $b$. %
Since $x$ always begins with an initial state declaration of the form ``\textit{the first beaker has [amount] [color], the second beaker has [amount] [color], ...}", tokens at position $8b-8\cdots 8b-1$ of $x$ correspond to the initial state of beaker $b$. (Each state has 8 tokens: `the', `$b$th', `be', `aker', `has', `[amount]', `[color]', `,'). Thus,
\begin{align*}
\begin{split}
    & \texttt{loc}(\texttt{has-$v$-$c$}(b), E(x)) = \\
    & \quad \texttt{mean}(E(x)[8b-8],\cdots, E(x)[8b-1])
\end{split}
\end{align*}

We train a linear probe $\texttt{cls}_\theta$ to predict the final beaker state given the encoded representation $E(x)$ of text $x$.
For our probe, we learn linear projection weights $W^{(d\times d)}$ and bias $b^{(d)}$ to maximize the dot product between the LM representation and the embedded proposition. Formally, it computes $v_b^{(max)}, c_b^{(max)}$ as
\begin{align}
\small
\begin{split}
& v_b^{(max)}, c_b^{(max)} = \arg\max_{v',c'}\big(\texttt{embed}(\texttt{has-$v'$-$c'$}(b))\cdot \\
& \quad (W\cdot \texttt{loc}(\texttt{has-$v'$-$c'$}(b),E(x))+b)\big)
\end{split}
\end{align}
In other words, $v_b^{(max)}, c_b^{(max)}$ are the values of $v$ and $c$ that maximize this dot product.
The probe then returns
\begin{align}
\small
\begin{split}
\texttt{cls}_{\theta}(\texttt{embed}( & \texttt{has-$v$-$c$}(b)), \texttt{loc}(\texttt{has-$v$-$c$}(b), E(x))) \\
& = \begin{cases}
T \text{ if } v,c = v_b^{(max)}, c_b^{(max)} \\
F \text{ if } v,c \neq v_b^{(max)}, c_b^{(max)}
\end{cases}
\end{split}
\label{eq:alchemy_probe}
\end{align}
Note that $\texttt{cls}_{\theta}$ selects the optimal final state \textit{per beaker}, from the set of all possible states of beaker $b$, taking advantage of the fact that only one proposition can be true per beaker.

\paragraph{Textworld Probe}
\label{sec:tw_probe}
For Textworld, the proposition embedder converts propositions $\phi$ to natural language descriptions %
$\widetilde{\phi}$ (``\texttt{the $o$ is $p$}" for properties and ``\texttt{the $o_1$ is $r$ $o_2$}" for relations) and encodes them with the BART or T5 LM encoder.

Given a proposition $p(o)$ pertaining to an entity or $r(o_1, o_2)$ pertaining to an entity pair, 
we define localizer $\texttt{loc}$ to map the proposition to tokens of $E(x)$ corresponding to \textit{all mentions} of its arguments, and averages across those tokens:
\begin{align}
\small
\begin{split}
    \texttt{all\_idx}(e) & = \text{set of indices of $x$ correspond} \\
    & \quad\,\,\text{-ing to all instances of $e$} \\
    \texttt{loc}(r(o_1,o_2),E(x)) & =
    \texttt{mean}\big(E(x)[ \texttt{all\_idx}(o_1)\cup \\
    & \qquad\qquad\qquad\,\,\,\texttt{all\_idx}(o_2)]\big) \\
    \texttt{loc}(p(o),E(x)) & = \texttt{mean}\left(E(x)[\texttt{all\_idx}(o)]\right)
\end{split}
\label{eq:tw_localizer}
\end{align}

We train a bilinear probe $\texttt{cls}_{\theta}$ that classifies each (proposition embedding, LM representation) pair to $\{\mathit{T, F, ?}\}$.
The probe has parameters $W^{(3\times d\times d)},b^{(3)}$ and performs the following bilinear operation:
\begin{align}
\small
\begin{split}
\texttt{scr}(\phi, E(x)) = \texttt{embed}(\phi)^T\cdot W\cdot \texttt{loc}(\phi, E(x)) + b
\end{split}
\end{align}
where $\texttt{scr}$ is a vector of size 3, with one score per $\mathit{T, F, ?}$ label. The probe then takes the highest-scoring label
\begin{align}
\small
\begin{split}
& \texttt{cls}_{\theta}(\texttt{embed}(\phi), \texttt{loc}(\phi, E(x))) = \\
& \begin{cases}
T \text{ if } \texttt{scr}(\phi, E(x))[0] > \texttt{scr}(\phi, E(x))[1], \texttt{scr}(\phi, E(x))[2] \\
F \text{ if } \texttt{scr}(\phi, E(x))[1] > \texttt{scr}(\phi, E(x))[2], \texttt{scr}(\phi, E(x))[0] \\
? \text{ if } \texttt{scr}(\phi, E(x))[2] > \texttt{scr}(\phi, E(x))[0], \texttt{scr}(\phi, E(x))[1]
\end{cases}
\end{split}
\label{eq:tw_probe}
\end{align}

\subsection{Localization Experiment Details + Additional Results (\S\ref{sec:localization})}
\label{sec:appendix_localization}

Below we provide a specific, formulaic account of each of our localizer experiments.

\paragraph{Mentions vs. Other Tokens (\S\ref{sec:which_token}) -- Alchemy}
Recall %
that we train a probe for each $(t,\Delta)$ pair 
to extract propositions about $b$ from token
$t_{b+\Delta}\in\texttt{toks}_{b+\Delta}$, where $\Delta$ is the beaker offset.
Specifically, the localizer for this probe
takes form
\begin{align*}
\small
\begin{split}
    & \texttt{off}: \\
    & \quad \{\textit{`the'}\to 0, \textit{[position]}\to 1, \textit{`be'}\to 2, \textit{`aker'}\to 3, \\
    & \quad \textit{`has'}\to 4, \textit{[amount]}\to 5, \textit{[color]}\to 6, \textit{`,'}\to 7 \}\\
    & \texttt{loc}_{(t,\Delta)}(\texttt{has-$v$-$c$}(b, E(x))) = \\
    & \quad\texttt{mean}(E(x)[8(b+\Delta)-8+\texttt{off}(t)])
\end{split}
\end{align*}

The full token-wise results for beaker states in a 3-beaker (24-token) window around the target beaker is shown in Figure~\ref{fig:alchemy_tokens_full} (Top/Middle).

Additional localizer ablations results for a BART probe trained and evaluated on synthetic Alchemy data are shown in Figures~\ref{fig:alchemy_tokens_full} (Bottom). %
Similar to the non-synthetic experiments, we point the localizer to just a single token of the initial state. %
Interestingly, BART's distribution looks very different in the synthetic setting. Though state information is still local to the initial state description of the target beaker, it is far more distributed within the description--concentrated in not just the amount and color tokens, but also the mention tokens.

\paragraph{Mentions vs. Other Tokens (\S\ref{sec:which_token}) -- Textworld}
The specific localizer for this experiment has form
\begin{align*}
\small
\begin{split}
    & \texttt{loc}(r(o_1,o_2),E(x)) = \\
    & \quad \texttt{mean}(E(x)[ \texttt{all\_idx}(\texttt{remap}(o_1)) \cup \\
    & \quad\qquad\qquad\quad\texttt{all\_idx}(\texttt{remap}(o_2))]) \\
    & \texttt{loc}(p(o), E(x)) = \texttt{mean}(E(x)[ \texttt{all\_idx}(\texttt{remap}(o))])
\end{split}
\end{align*}
Note the evaluation set for this experiment is slightly different as we exclude contexts which do not mention $\texttt{remap}(w)$. 

\begin{table}[]
    \centering
    \small
    \begin{tabular}{rcc}
    \toprule
        & \multicolumn{2}{c}{\textbf{Alchemy}} \\%&
    \cmidrule(lr){2-3}
        & Entity EM & State EM \\ 
    \midrule
        main probe (\S\ref{sec:probe}) & 75.0 & 7.55 \\ 
    \midrule
        human-grounded-features (\S\ref{sec:appendix_embed_ablate}) 
        & 45.7 & 0.71 \\ %
        synthetic data & 88.2 & 35.9 \\%& - & - \\
    \bottomrule
    \end{tabular}
    \vspace{-.3em}
    \caption{Additional Alchemy results. We compare our full encoded-NL embedder with a featurized embedder (\S\ref{sec:appendix_embed_ablate}). 
    We also report results on synthetic data.}.
    \label{tab:alchemy_appendix}
    \vspace{-1em}
\end{table}
\paragraph{Which Mention? (\S\ref{sec:which_mention}) -- first/last}
The localizer for this experiment is constructed by replacing all instances of $\texttt{all\_idx}$ in Eq.~\ref{eq:alchemy_probe} with either $\texttt{first\_idx}$ or $\texttt{last\_idx}$, defined as:
\begin{align*}
\small
\begin{split}
    \texttt{first\_idx}(e) & = \text{set of indices of $x$ correspond-} \\
    & \quad\,\,\text{ing to first instance of $e$} \\
    \texttt{last\_idx}(e) & = \text{set of indices of $x$ correspond-} \\
    & \quad\,\,\text{ing to last instance of $e$} \\
\end{split}
\end{align*}

\paragraph{Which Mention? (\S\ref{sec:which_mention}) -- single- vs. both-entity probe.}
The specific localizer for the single-entity probe has form
\begin{align*}
\small
\begin{split}
    \texttt{loc}(r(o_1,o_2),E(x)) & =
    \begin{Bmatrix}
    \texttt{mean}(E(x)[\texttt{all\_idx}(o_1)]), \\
    \texttt{mean}(E(x)[\texttt{all\_idx}(o_2)]) \\
    \end{Bmatrix} \\
    \texttt{loc}(p(o), E(x)) & = \texttt{mean}(E(x)[ \texttt{all\_idx}(o)])
\end{split}
\end{align*}
Note the localizer returns a 2-element \textit{set} of encodings from each relation.
We train the probe to decode $r(o_1,o_2)$ from \textit{both} elements of this set.

The full results are in Table~\ref{tab:textworld_simple_relations}. 
As shown, the both-mentions probe is slightly better at both decoding relations and properties.
However, this may simply be due to having less candidate propositions per entity pair, than per entity (which includes relations from \textit{every other} entity paired with this entity).
For example, entity pair \textit{(apple, chest)} has only three possibilities: \texttt{in(apple, chest)} is True/Unknown/False, while singular entity \textit{(chest)} has much more: \texttt{in(apple, chest)}, \texttt{in(key, chest)}, \texttt{open(chest)}, etc. can each be True/Unknown/False.
A full set of results broken down by property/relation can be found in Table~\ref{tab:probe_appendix}.
Overall, the single-entity probe outperforms all baselines, suggesting that each entity encoding contains information about its relation with other entities.

\subsection{Proposition Embedder Ablations}
\label{sec:appendix_embed_ablate}
We experiment with a featurized $\texttt{embed}$ function in the Alchemy domain.
Recall from Section~\ref{sec:probe} and~\ref{sec:appendix_probe_details} that our main probe uses encoded natural-language assertions of the state of each beaker (Eq.~\ref{eq:embed}).
We experiment with featurized vector where each beaker proposition is the concatenation of a 1-hot vector for beaker position and a sparse vector encoding the amount of each color in the beaker (with 1 position per color).
For example, if there are $2$ beakers and $3$ colors $[\texttt{green,red,brown}]$, \texttt{has-3-red(2)} is represented as $[0,1,0,3,0]$. A multi-layer perceptron is used as the $\texttt{embed}$ function to map this featurized representation into a dense vector, which is used in the probe as described by Eq.~\ref{eq:alchemy_probe}. In this setting, the $\texttt{embed}$ MLP is optimized jointly with the probe.

Results are shown in Table~\ref{tab:alchemy_appendix}. Using a featurized representation (45.7) is significantly worse than using an encoded natural-language representation (75.0), suggesting that the form of the fact embedding function is important. In particular, \textit{the encoding is linear in sentence-embedding space, but nonlinear in human-grounded-feature space.}

\subsection{Error Analysis}
\label{sec:appendix_error_analysis}

\begin{table*}[t!]
    \centering
    \small
    \begin{tabular}{rccccccccc}
    \toprule
        & \textbf{Overall} & \textbf{Relations} & \textbf{Properties} & \multicolumn{3}{c}{\textbf{True Facts}} & \multicolumn{3}{c}{\textbf{False Facts}}  \\
    \cmidrule(lr){2-2}\cmidrule(lr){3-3}\cmidrule(lr){4-4}\cmidrule(lr){5-7}\cmidrule(lr){8-10}
        & EM & EM & EM & Pre & Rec & F1 & Pre & Rec & F1 \\
    \midrule
        & 48.7 & 49.6 & 94.5 & 95.1 & 98.1 & 96.4 & 99.6 & \textbf{98.9} & 99.2 \\
    \midrule
        +pretrain, -finetune & 23.2 & 32.7 & 75.4 & 93.2 & 94.9 & 93.8 & 97.0 & 96.1 & 96.4 \\
        -pretrain, +finetune & 14.4 & 26.8 & 44.0 & 87.3 & 86.6 & 86.3 & 93.3 & 93.0 & 92.9 \\
        randomly initialized & 11.3 & 21.9 & 35.4 & 91.5 & 83.8 & 87.2 & 91.8 & 84.1 & 86.5 \\
        no LM & 1.77 & 24.8 & 33.4 & 88.3 & 80.9 & 84.4 & 88.8 & 86.9 & 87.6 \\
        no change & 9.73 & 30.1 & 9.73 & 77.9 & 73.0 & 75.3 & 79.1 & 61.8 & 68.9 \\
    \midrule
        locality (first) & 49.6 & 50.9 & 88.5 & 95.4 & 97.2 & 96.1 & 99.1 & 98.7 & 98.9 \\
        locality (last) & \textbf{55.1} & \textbf{56.6} & \textbf{96.5} & \textbf{96.1} & \textbf{98.7} & \textbf{97.3} & \textbf{99.7} & 98.9 & \textbf{99.3} \\
    \bottomrule
    \end{tabular}
    \vspace{-.3em}
    \caption{TextWorld Probing. Metrics are reported on \textit{whole state}. Precision is computed against \textit{all} gold, ground-truth True facts in the state. Recall is computed against the \texttt{label}-model-generated True facts in the state. All numbers reported are averaged across all samples.
    Relations are overall much harder to probe than properties.
    }
    \label{tab:probe_appendix}
\end{table*}
We run error analysis on the BART model.
For the analysis below, it is important to note that we make no distinction between \textit{probe errors} and \textit{representation errors}---we do not know whether the errors are attributable to the linear probe's lack of expressive power, or whether the underlying LM indeed does fail to capture certain phenomena.
We note that a \textit{BART decoder} trained to decode the final information state from $E(x)$ is able to achieve 53.5\% state EM on Alchemy (compared to 0\% on random initialization baseline) whereas the linear decoder was only able to achieve 7.55\% state EM---suggesting that certain state information may be non-linearly encoded in NLMs.

\paragraph{Alchemy}
The average number of incorrect beakers per sample is 25.0\% (2.7 beakers out of 7). 

We note that the distribution is skewed towards longer sequences of actions, where
the \% of wrong beakers increases from 11.3\% (at 1 action) to 24.7\% (2 actions), 30.4\% (3 actions), 33.4\% (at 4 actions).
For beakers not acted upon (final state unchanged), the error rate is 13.3\%. 
For beakers acted upon (final state changed), the error rate is 44.6\%.
Thus, errors are largely attributed to failures in reasoning about the effects of actions, rather than failures in decoding the initial state. (This is unsurprising, as in Alchemy, the initial state is explicitly written in the text---and we're decoding from those tokens).

For beakers that were predicted wrong, 36.8\% were predicted to be its unchanged initial state and the remaining 63.2\% were predicted to be empty --- thus, probe mistakes are largely attributable to a tendency to over-predict empty beakers. 
This suggests that
the downstream decoder may tend to generate actions too conservatively (as empty beakers cannot be acted upon). Correcting this could encourage LM generation diversity.

Finally, we examine what \textit{type} of action tends to throw off the probe.
When there is a \textit{pouring} or \textit{mixing}-type action present in the sequence, the model tends to do worse (25.3\% error rate for drain-type vs. 31.4 and 33.3\% for pour- and mix-type), though this is partially due to the higher concentration of drain actions in short action sequences.

\paragraph{Textworld}
Textworld results, broken down by properties/relations, are reported in Table~\ref{tab:probe_appendix}.
The probe seems to be especially bad at classifying relations, which make sense as relations are often expressed indirectly. A breakdown of error rate for each proposition type is shown in Table~\ref{tab:tw_error_breakdown}, where we report what \% of that type of proposition was labelled incorrectly, each time it appeared.
This table suggests that the probe consistently fails at decoding locational relations, i.e. failing to classify \texttt{east-of(kitchen,bedroom)} and \texttt{west-of(kitchen,bedroom)} as True, %
despite the layout of the simple domain being fixed. %
One hypothesis is that location information is made much less explicit in the text, and usually require reasoning across longer contexts and action sequences. For example, classifying \texttt{in(key, drawer)} as True simply requires looking at a single action: \textit{$>$ put key in drawer}. However, classifying \texttt{east-of(kitchen,bedroom)} as True requires reasoning across the following context:
\begin{quote}
  \textit{You are in the bedroom [\dots] \\
$>$ go east \\
You enter the kitchen.}
\end{quote}
where the ellipses possibly encompass a long sequence of other actions.

\begin{table}[]
    \centering
    \begin{tabular}{rl}
    \toprule
        Proposition Type & Error Rate \\
    \midrule
        \texttt{\{north|south|east|west\}} & \multirow{2}{*}{11.8\%} \\
        \texttt{-of(A,B)} \\
        \texttt{is-on(A,B)} & 6.17\% \\
        \texttt{is-in(A,B)} & 1.20\% \\
        \texttt{locked(A)} & 0.47\% \\
        \texttt{closed(A)} & 0.35\% \\
        \texttt{eaten(A)} & 0.049\% \\
        \texttt{open(A)} & 0.039\% \\
    \bottomrule
    \end{tabular}
    \caption{Error rate per proposition type in Textworld.}
    \label{tab:tw_error_breakdown}
\end{table}

\subsection{Infrastructure and Reproducibility}

We run all experiments on a single 32GB NVIDIA Tesla V100 GPU.
On both Alchemy and Textworld, we train the language models to convergence, then train the probe for 20 epochs.
In Alchemy and Textworld, both training the language model and the probe takes approximately a few (less than 5) hours.
We probe BART-base, a 12-layer encoder-decoder Transformer model with 139M parameters, and T5-base, a 24-layer encoder-decoder Transformer model which has 220M parameters. Our probe itself is a linear model, with only two parameters (weights and bias).

\end{document}